\documentclass[10pt,twocolumn,letterpaper,table]{article}
\usepackage{multirow} 
\usepackage{wacv}
\usepackage{times}
\usepackage{epsfig}
\usepackage{graphicx}
\usepackage{amsmath}
\usepackage{amssymb}
\usepackage{hyperref}
\usepackage{textcomp}
\usepackage{makecell}
\usepackage[dvipsnames]{xcolor}
\usepackage{booktabs} 
\usepackage{siunitx} 
\usepackage{color,soul}
\usepackage[normalem]{ulem}
\usepackage{subcaption} 

\wacvfinalcopy 

\def\wacvPaperID{****} 

\ifwacvfinal\fi
\definecolor{yolo}{RGB}{175, 98, 31}
\definecolor{mask-rcnn}{RGB}{31, 50, 175}
\definecolor{ground-truth}{RGB}{120,240,120}
\definecolor{ours}{RGB}{221, 169, 55}
\definecolor{cv}{RGB}{66, 149, 244}
\begin{document}

\title{Multiple Object Forecasting: Predicting Future Object Locations in Diverse Environments}

\author{Olly Styles \hspace{2.4cm} Tanaya Guha \hspace{2cm} Victor Sanchez\\University of Warwick\\
{\tt\small \{o.c.styles, tanaya.guha, v.f.sanchez-silva\}@warwick.ac.uk}
}

\maketitle
\ifwacvfinal\thispagestyle{empty}\fi

\begin{abstract}
    This paper introduces the problem of multiple object forecasting (MOF), in which the goal is to predict future bounding boxes of tracked objects. In contrast to existing works on object trajectory forecasting which primarily consider the problem from a birds-eye perspective, we formulate the problem from an object-level perspective and call for the prediction of full object bounding boxes, rather than trajectories alone. Towards solving this task, we introduce the Citywalks dataset, which consists of over 200k high-resolution video frames. Citywalks comprises of footage recorded in 21 cities from 10 European countries in a variety of weather conditions and over 3.5k unique pedestrian trajectories. For evaluation, we adapt existing trajectory forecasting methods for MOF and confirm cross-dataset generalizability on the MOT-17 dataset without fine-tuning.  Finally,  we present STED, a novel encoder-decoder architecture for MOF. STED combines visual and temporal features to model both object-motion and ego-motion, and outperforms existing approaches for MOF. Code \& dataset link: \url{https://github.com/olly-styles/Multiple-Object-Forecasting}
\end{abstract}
\vspace{-0.8cm}

\section{Introduction}

Predicting future events in video is a core problem in computer vision that has been studied in several contexts such as human action prediction \cite{actionprediction},  semantic forecasting \cite{lecun-semantic}, and road agent trajectory forecasting \cite{desire}. In this work, we focus on the task of pedestrian trajectory forecasting from video data, which has seen considerable research attention over recent years \cite{activityforecasting,goaldirected,sociallstm,socialgan,behavior-cnn,fpl}. Humans are a particularly challenging class of objects to predict, as they exhibit highly dynamic motion and may change speed or direction rapidly.
\begin{figure}[t]
\begin{center}
   \includegraphics[width=0.95\linewidth]{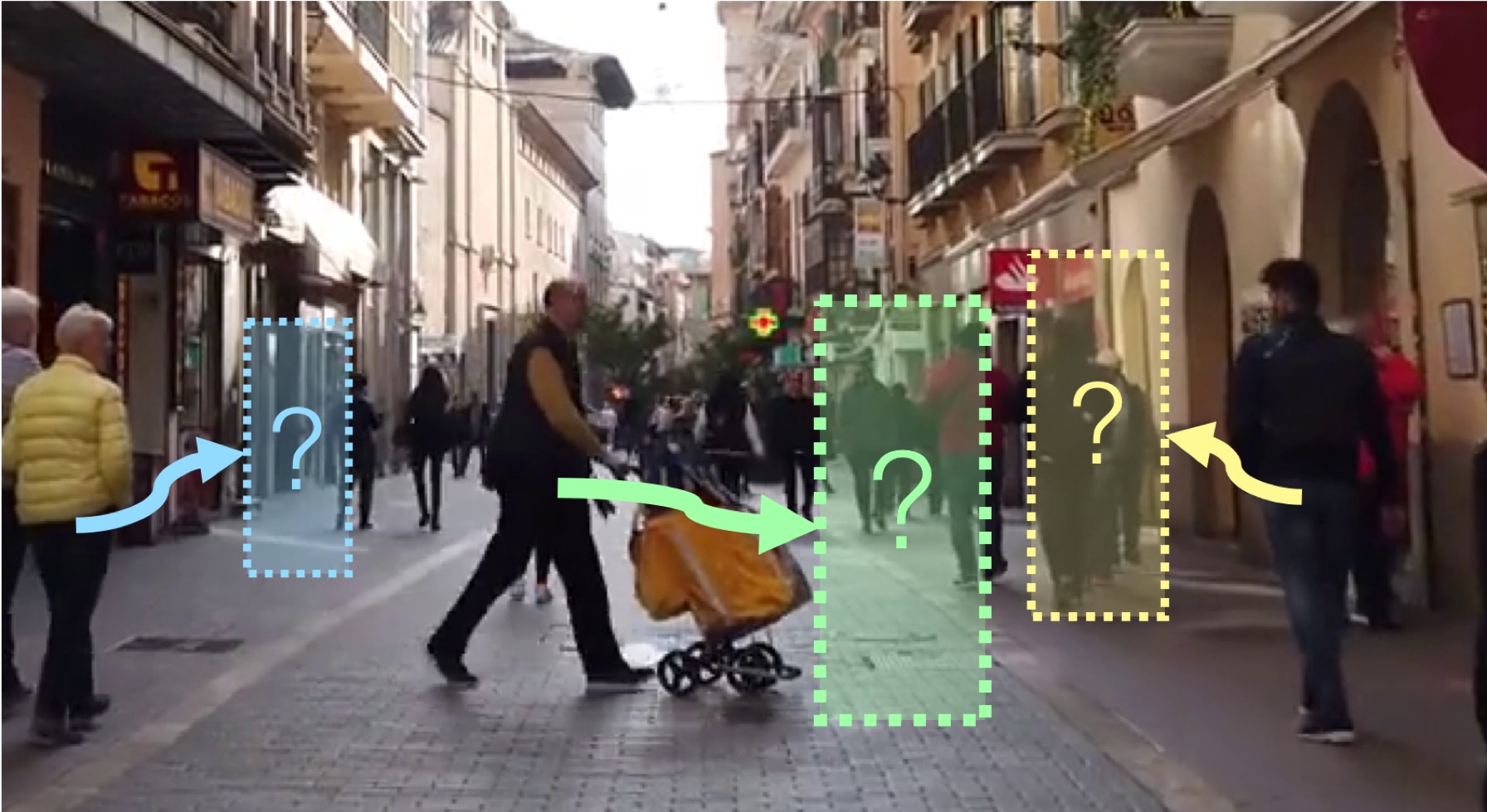}
\end{center}
\vspace{-0.4cm}
   \caption{We introduce the new task of multiple object forecasting and the Citywalks dataset to facilitate future research.}
 \vspace{-0.2cm}
\label{fig:pull}
\end{figure}

Much of the existing work on pedestrian trajectory forecasting considers the problem from a birds-eye view using footage from a fixed overhead camera, often considering each pedestrian as a single point in space \cite{sociallstm,socialgan,behavior-cnn}. This setting is effective for modeling crowd motion patterns and interactions with the environment. However, by simplifying each pedestrian as a point in space, salient visual features such as person appearance, body language, and individual characteristics are not considered. Prior research has shown that these features are of importance for trajectory prediction in settings such as anticipating if a pedestrian will cross the road \cite{atthekerb,agreeing}. Furthermore, overhead perspectives are often not available in practical applications. As a result, trajectory forecasting from an object-level perspective has been studied in recent years \cite{fpl}, although suffers from a lack of large, high-quality datasets and standardized evaluation protocols.

Motivated by the above observations, we introduce a new formalization of the trajectory forecasting task: multiple object forecasting (MOF) (Fig. \ref{fig:pull}). MOF follows the same formulation as the popular multiple object tracking (MOT) task, but rather is concerned with predicting \textit{future} object bounding boxes and tracks in upcoming video frames, rather than the bounding boxes and tracks in the current frame. Future bounding box prediction has previously been studied in constrained settings such as on-board a moving vehicle with odometry information \cite{longterm,egocentric}. In contrast, MOF follows the unconstrained MOT setting, which utilizes only image information where data from other sensors is not available. This setup poses several challenges, such as variations in object scale, non-linear motions, and ego-motion. MOF has a number of possible applications such as object tracking \cite{tracking-by-prediction} (particularly through occlusions), robotic navigation \cite{robotanticipation}, and autonomous driving \cite{willthepedcross}.

To facilitate research on the MOF problem, we construct the Citywalks dataset. Citywalks is a large and diverse dataset collected from a first-person perspective in 21 European cities with considerable variability in many facets such as weather, object appearance, illumination, object scale, and pedestrian density. Citywalks is annotated using automated methods for detection and tracking and is considerably more diverse than existing datasets \cite{eth,stanforddronedataset,mot-16} for trajectory forecasting. We evaluate existing models adapted for MOF on Citywalks and propose a novel encoder-decoder model. Our model, STED,  combines visual features extracted from optical flow with temporal features and outperforms existing models on the MOF task.

The contributions of this work are as follows:

\begin{enumerate}
\vspace{-0.2cm}
    \item We introduce MOF, a new formulation of the trajectory forecasting problem (Section \ref{sec:MOF}).
\vspace{-0.2cm}
\item We introduce and publicly release Citywalks, a challenging dataset for MOF with considerably more geographical variety than existing datasets (Section \ref{sec:dataset}).
\vspace{-0.2cm}
\item We propose STED, a \underline{S}patio-\underline{T}emporal \underline{E}ncoder-\underline{D}ecoder model for MOF which combines visual and temporal features (Section \ref{sec:model}). Experimental evaluation using two datasets confirms the benefits of our proposed approach (Section \ref{sec:perf-eval}).
\end{enumerate}

\section{Related work}
In this section, we summarize the main contributions in the fields of pedestrian trajectory forecasting and MOT. We also provide an overview of existing datasets for both tasks and their limitations.

\subsection{Multiple object tracking}

Methods for MOT typically follow a tracking-by-detection paradigm that relies heavily on the accuracy of single-frame detections and models to associate detections across time. Reasonable MOT performance can be obtained with high-quality detections and simple constant velocity motion assumptions \cite{sort}, and better still when combined with a visual appearance association metric \cite{deepsort}. Constructing more sophisticated methods capable of modeling non-linear motion can improve tracking performance, particularly in scenarios with occlusion \cite{tracking-by-prediction}. However, trajectory forecasting for improved tracking is challenging due to small datasets, which results in overfitting. One approach proposed to overcome this issue is to consider the future trajectory as a binary classification problem \cite{tracking-untrackable} or using explicit external memory to avoid memorization \cite{recurrentautoregressive}. We adopt a more straightforward approach to address overfitting: building a larger dataset.

\subsection{Pedestrian trajectory forecasting}

Pedestrian trajectory forecasting has been studied extensively in a surveillance setting from fixed cameras from a birds-eye view  \cite{sociallstm,behavior-cnn,socialgan,lookingtorelations,ouyang-trajectory-prediction}. Methods typically focus on interactions between pedestrians and social conventions such as the pioneering Social Long-Short-Term-Memory (Social-LSTM) model \cite{sociallstm}, in addition to scene semantics. These methods do not typically consider visual cues, and many simplify each pedestrian to a point in space. Recently, Liang et al. \cite{peeking} proposed one of the first approaches for trajectory forecasting using visual features. Their method encodes appearance using a person keypoint detector and joint modeling of future pedestrian trajectory and activity.

Most related to our paper, a small number of works consider trajectory forecasting from an object-level perspective. Predicting object trajectories from on-board moving vehicles, in particular, has been studied extensively \cite{willthepedcross, longterm, dtp}. Methods typically use additional information sources specific to a vehicle setting, such as odometry information. In an inspiring work outside of the vehicle domain, Yagi et al. \cite{fpl} propose a model that uses past locations, ego-motion, and pedestrian keypoints to estimate future trajectory in first-person videos. Their model outperforms existing state-of-the-art approaches; however, accurate pedestrian keypoint estimation is not always practical, especially in low-resolution or low-lighting scenarios. In contrast, our approach does not rely on pedestrian keypoint estimation.

\begin{figure*}
\begin{center}
\includegraphics[width=0.95\linewidth]{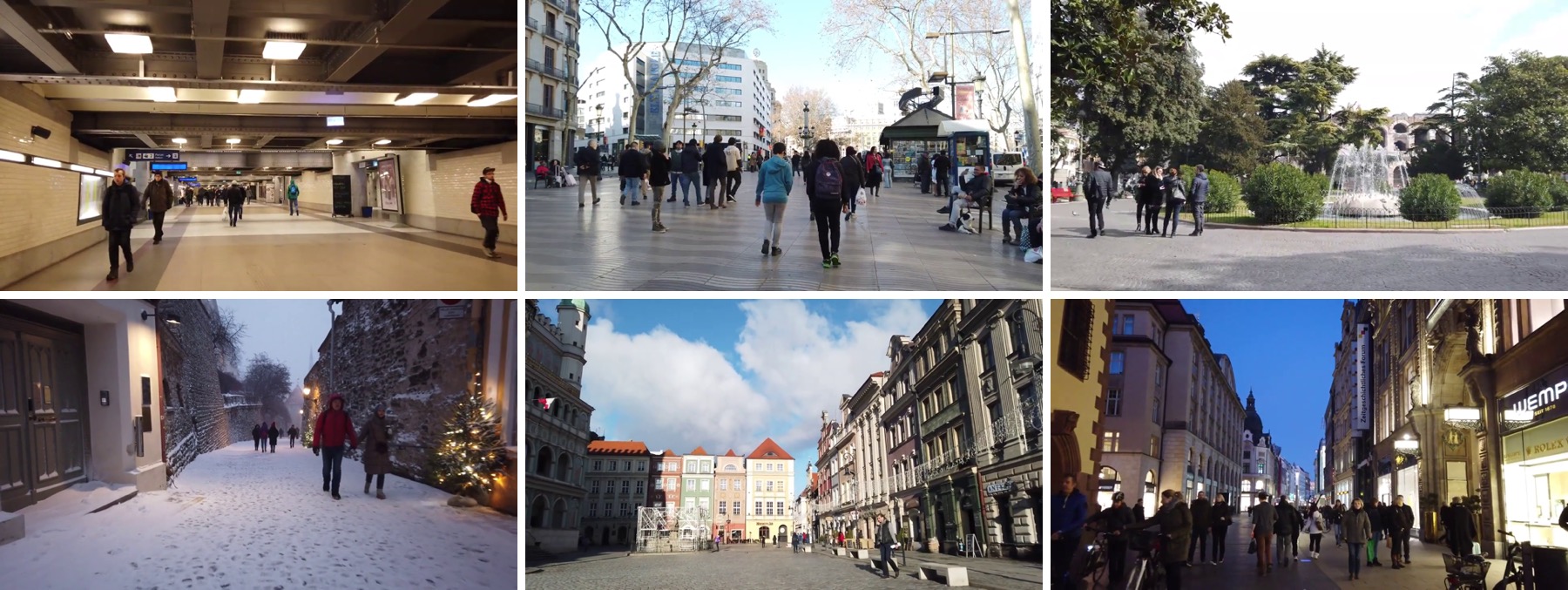}
\end{center}
\vspace{-0.3cm}
   \caption{Example frames from the Citywalks dataset. Citywalks is markedly larger and more diverse than existing datasets.}
\label{fig:mof-frames}
\vspace{-0.1cm}
\end{figure*}

\subsection{Existing datasets}

Many large datasets with annotated pedestrian bounding boxes have been released such as Citypersons \cite{citypersons}, BDD-100K \cite{bdd100k} and EuroCity Persons \cite{eurocitypersons}. However, these datasets do not contain object tacking annotations. Older datasets such are KITTI \cite{kitti} and Caltech-USA \cite{caltech-usa} provide full object tracks, although these datasets are considerably smaller with more limited geographical variety than our new dataset.

Several datasets have been created explicitly for pedestrian trajectory forecasting, such as UCY \cite{ucy}, ETH \cite{eth}, and Stanford Drone \cite{stanford-drone}. These datasets are recorded from a birds-eye view, making them suitable for modeling social and environmental factors. However, such datasets are not well suited to MOF due to being captured at a perspective from which extracting visual features is challenging.

Few public datasets exist for object-level view trajectory forecasting. Most similar to ours, the MOT-17 dataset \cite{mot-16} contains annotated pedestrian bounding boxes from both first-person and overhead cameras. However, MOT-17 contains only 14 video sequences. Our dataset, Citywalks, contains 358 video sequences.

\section{Multiple object forecasting} \label{sec:MOF}

MOF follows a similar problem formulation to the prevalent MOT task. In this section, we formalize MOF and the metrics used for evaluating models.

\subsection{Problem formulation}  \label{sec:problem}
Consider a sequence of $n$ video frames $f_0,f_1,\ldots,f_{n-1}$. Given the $t^{th}$ frame $f_t$, the task of object detection is to associate each identifiable object $i\in \mathcal{I}$ in the frame with a set of coordinates $b^i_t=(x_t,y_t,w_t,h_t)$ which represent the centroid $(x_t,y_t)$, width, and height of the object bounding box, and $\mathcal{I}$ is the set of all identifiable objects. Given all the framewise detections $\{b^i_0\}, \{b^i_1\},\ldots,\{b^i_n\}$ for all $i\in\mathcal{I}$, the task of MOT is to associate each detection $b^i_t$ with a unique object identifier $k\in {1,2\ldots K }$, where $K$ is the total number of unique objects across all frames, such that each object is tracked across the set of $n$ frames.

We extend the MOT task to MOF, shown in Fig. \ref{fig:pull}. Given $f_{t-p},f_{t-p+1},\ldots,f_t$ with associated object detections $\{b^i_{t-p}\}, \{b^i_{t-p+1}\}\ldots\{b^i_t\}$ and tracks, we define MOF as the joint problem of predicting the future bounding boxes $\{b^i_{t+1}\}, \{b^i_{t+2}\},\ldots,\{b^i_{t+q}\}$ and associated object tracks of the upcoming $f_{t+1},f_{t+2},\ldots,f_{t+q}$ video frames for each object present in frame $f_t$, where $p$ is the number of past frames used as input and $q$ is the number of future frames to be predicted. In this work, we use $p=30$ and $q=60$, corresponding to 1 second in the past and 2 seconds into the future at 30Hz.

\subsection{Evaluation metrics} \label{sec:metrics}

We adopt the average displacement error (ADE) and final displacement error (FDE) metrics from the trajectory forecasting literature \cite{sociallstm}. ADE is defined as the mean Euclidean distance between predicted and ground-truth bounding box centroids for all predicted bounding boxes, and FDE is defined similarly for the centroid at the final timestep only. We also use the average and final intersection-over-union (AIOU and FIOU) metrics. AIOU is defined as the mean IOU of the predicted and ground truth bounding boxes for all predicted boxes, and FIOU is the IOU for the box at the final timestep only.

\section{Citywalks Dataset} \label{sec:dataset}

Our newly-constructed Citywalks dataset comprises of 358 video sequences containing footage from 21 different cities in 10 European countries.

\begin{figure*}[t]
     \centering
     \begin{subfigure}[t]{0.33\textwidth}
         \centering
         \includegraphics[width=\textwidth]{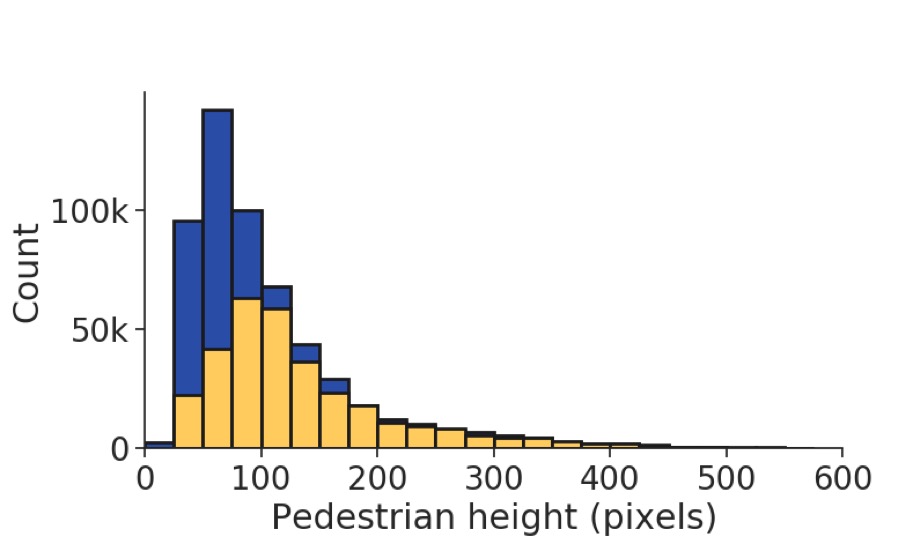}
     \end{subfigure}
     \begin{subfigure}[t]{0.33\textwidth}
         \centering
         \includegraphics[width=\textwidth]{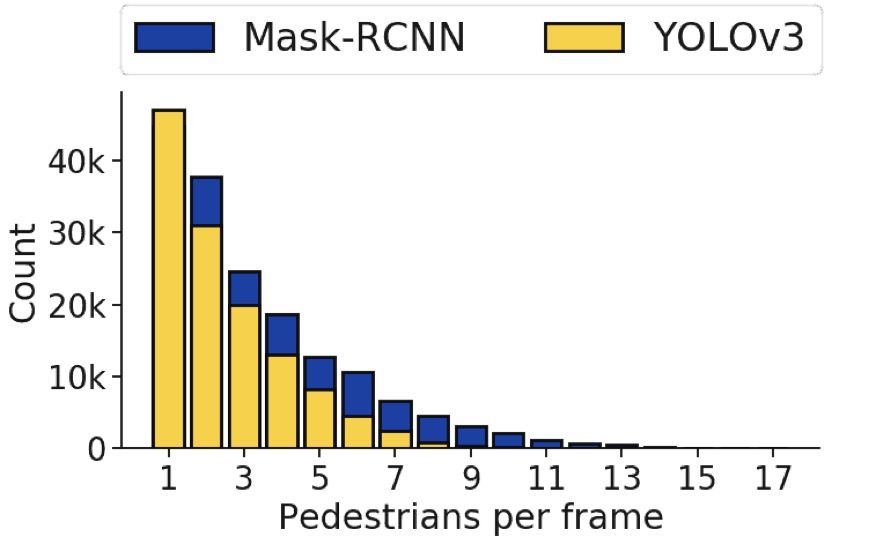}
     \end{subfigure}
     \begin{subfigure}[t]{0.33\textwidth}
         \centering
         \includegraphics[width=\textwidth]{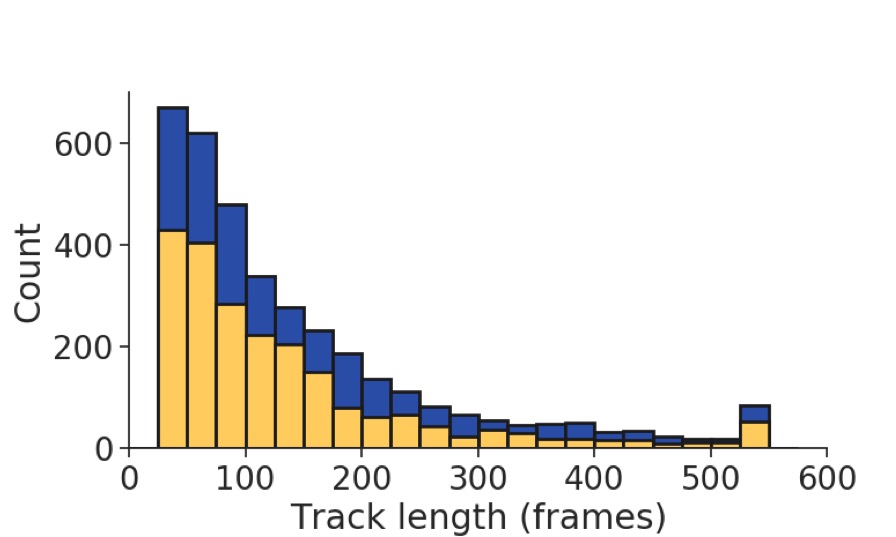}
     \end{subfigure}
        \caption{Citywalks annotation statistics.}
        \label{fig:annotation-stats}
\end{figure*}

\subsection{Data collection}
We extract footage from the online video-sharing site YouTube\footnote{Videos are obtained from \url{https://www.youtube.com/c/poptravelorg}}. Each original video consists of first-person footage recorded using an Osmo Pocket camera with gimbal stabilizer held by a pedestrian walking in one of the many environments for between 50 and 100 minutes. Videos are recorded in a variety of weather conditions, as well as both indoor and outdoor scenes. Example frames showcasing the variety of the dataset are shown in Fig. \ref{fig:mof-frames}.

\newcolumntype{C}[1]{>{\centering\arraybackslash}p{#1}}

\begingroup
\begin{table}[h]
\begin{center}   \caption{Citywalks metadata.}
  \vspace{-0.1cm}
  \begin{tabular}{ C{4.15cm}  C{3.25cm} }
    \toprule
    Video clips & 358  \\[1pt]
    Resolution & $1280 \times 720$ \\[1pt]
    Framerate & 30hz \\[1pt]
    Clip length & 20 seconds \\[1pt]
    Unique cities & 21 \\[1pt]
    Weather conditions & Sun/Rain/Snow/Overcast  \\[1pt]
    Time of day &  Day/Night  \\[1pt]
    Labelled objects per frame &  0 - 17  \\[1pt]
    Unique tracks (YOLOv3) & 2201 \\[1pt]
    Unique tracks (Mask-RCNN) & 3623 \\[1pt]
    \bottomrule

  \end{tabular}\label{Citywalks-stats}
  \end{center}
  \vspace{-0.3cm}
  \end{table}
\endgroup

\subsection{Video clip filtering}
One of the fundamental challenges of MOF is the bounding box motion caused by both ego-motion and object motion. Large displacements resulting from significant ego-motion pose a problem and may overwhelm the training process. To mitigate the impact of large ego-motions, we filter the dataset by removing high motion segments. Global motion is estimated by extracting dense optical flow and selecting short video clips from windows with a mean optical flow magnitude below a threshold. Specifically, we downsample video frames to $128 \times 64$ pixels for faster computation and extract dense optical flow using FlowNet2-S \cite{flownet2}. We then select 20-second clips from longer videos using segments containing frames that do not exceed a mean optical flow magnitude threshold of 1.5.

\subsection{Annotations}
Once clips are selected, pedestrians are detected using an object detection algorithm. We provide annotations for two object detectors: YOLOv3 \cite{yolo3} and Mask-RCNN \cite{mask-rcnn}. Both detectors are trained using the MS-COCO \cite{coco} dataset and generalize well to Citywalks. For the YOLOv3 annotations, images are downsampled to $416 \times 416$ pixels before detection, to simulate detection quality under low processing time requirements. We use a resolution of $1024 \times 1024$ for detection using Mask-RCNN to obtain the best detection performance. Note that we leave any attempts to combine the two annotation sets (such as in \cite{strongandweak}) for future work. Following the detection phase, pedestrians are tracked using DeepSORT \cite{deepsort}, which uses a Kalman filter and person re-identification model to associate detections across frames. We then discard tracks shorter than 3 seconds as the previous 1 second of bounding box data is used to predict the next 2 seconds. Dropping short tracks reduces the number of false positives in the annotation set, as we observe that erroneous tracks typically do not last longer than 3 seconds. Each video clip is also manually annotated with the city of recording, time of day, and weather condition. Annotation statistics are shown in Fig. \ref{fig:annotation-stats}, and metadata are shown in Table \ref{Citywalks-stats}.

\vspace{-0.2cm}
\section{Proposed model} \label{sec:model}
\begin{figure*}[t]
\begin{center}
\includegraphics[width=\linewidth]{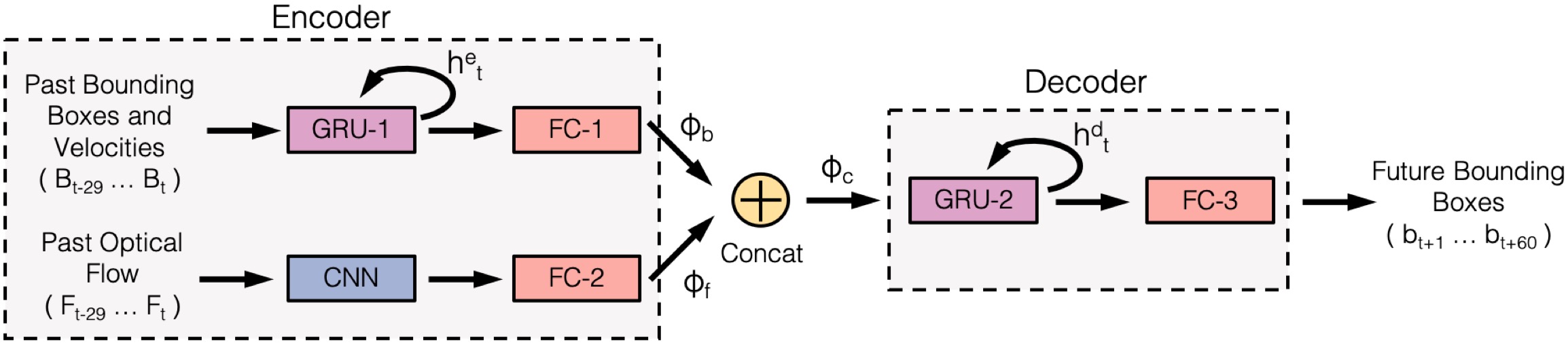}
\end{center}
    \vspace{-0.3cm}
   \caption{STED consists of a Gated Recurrent Unit (GRU), a Convolutional Neural Network (CNN), and two fully-connected (FC) layers for feature encoding. Our decoder takes the encoded feature vector $\phi_c$ as input and outputs predicted object bounding boxes for the next 2 seconds using another GRU and FC layer.}
\label{fig:ourmodel}
\end{figure*}
In this section, we present STED, an encoder-decoder architecture for MOF that combines visual and temporal features. The proposed architecture has three components: (i) A bounding box feature encoder based on a Gated Recurrent Unit (GRU) \cite{gru} that extracts temporal features from past object bounding boxes (ii) A CNN-based encoder that extracts motion features directly from optical flow, and (iii) a decoder implemented as another GRU for generating future bounding box predictions given the learned features. An overview of our model is shown in Fig. \ref{fig:ourmodel}.

\vspace{-0.1cm}

\subsection{Bounding box feature encoder} \label{sec:bb-encoder}

Our bounding box encoder extracts features from past bounding box coordinates of each object $i$ represented in terms of its centroid, width and height $b^i_t = (x_t,y_t,w_t,h_t)$. In addition, we compute the velocity in the $x$ and $y$ directions, $(v^x_t,v^y_t)$, change in width, $\Delta w_t$, and change in height, $\Delta h_t$. This results in an $8$-dimensional vector associated with each object bounding box  $B^i_t = (x_t,y_t,w_t,h_t,v^x_t,v^y_t, \Delta w_t, \Delta h_t)$.

For each observed timestep, a GRU (GRU-1 in Fig. \ref{fig:ourmodel}) takes the vector $B_t^i$ as input and outputs an updated hidden state vector $h^e_t$. This update is repeated for all timesteps, resulting in a single hidden state vector $h^e_t$ at the final timestep which summarizes the entire sequence of bounding boxes.  The 256-dimensional feature vector $\phi_b$ from a fully connected layer (FC-1 in Fig. \ref{fig:ourmodel}) is used as a compact representation of the history of bounding boxes.

\subsection{Optical flow feature encoder}
We adapt Dynamic Trajectory Predictor (DTP) \cite{dtp} to learn features directly from optical flow. Flow frames, $F_t$, are extracted from within object bounding boxes obtained using YOLOv3 or Mask-RCNN at each timestep. A stack of $10$ frames are sampled uniformly from timesteps $t-29$ to $t$ inclusively, representing 1 second of motion history. The stack of 10 horizontal and 10 vertical frames are used as input to a CNN which takes the $20 \times 224 \times 224$ stack of frames as input and is trained to predict future object bounding boxes. The $2048$-dimensional feature vector $\phi_f$ from the final fully connected layer (FC-2 in Fig. \ref{fig:ourmodel}) is used as a compact representation of optical flow features. As optical flow captures both object motion and ego-motion, the vector $\phi_f$ encodes information from these two motion sources. Using optical flow as the input of our encoder rather than features from a person keypoint estimation model \cite{fpl} avoids the challenges relating to inaccurate keypoint estimations.

\subsection{Decoder} \label{sec:decoder}
Following the feature encoding stage, we use another GRU to generate the estimated sequence of future bounding boxes, enabling the model to generate predictions for an arbitrary number of timesteps into the future. The two feature vectors, $\phi_f$ and $\phi_b$, are concatenated resulting in a single feature vector $\phi_c$ representing both optical flow and bounding box history. For each future timestep to be predicted, the decoder GRU (GRU-2 in Fig. \ref{fig:ourmodel}) receives two inputs: The concatenated feature vector $\phi_c$, and the internal hidden state $h^d_{t-1}$. The GRU outputs a new value for $h^d_t$ at each timestep. Given each generated hidden state, a final fully connected layer generates the predicted bounding box for each timestep. Rather than representing object bounding boxes by their absolute location  \cite{behavior-cnn} or relative displacement from the previous bounding box \cite{fpl}, we adopt the formulation of \cite{dtp} and represent the bounding box centroid as the relative change in velocity. The decoder generates a vector $(\Delta v^x,\Delta v^y, \Delta w,\Delta h)$, representing the change in velocity along the $x$ and $y$-axes, and the change in bounding box width and height. The untrained model is initialized to the case where $\Delta v^x = \Delta v^y = 0$ (constant velocity) and $\Delta w = \Delta h = 0$ (constant scale). This formulation results in a better initialization than absolute or relative locations.

\section{Performance evaluation} \label{sec:perf-eval}

\subsection{Baseline models}

\begin{table*}[t]
\setlength{\tabcolsep}{1pt} 
\begin{center}   \caption{Results averaged over 3 train-test splits on Citywalks with our two annotation sets using YOLOv3 and Mask-RCNN. DTP and FPL predict object centroids only, so IOU metrics are not applicable.}
\vspace{-0.2cm}
\label{tab:Citywalks-results}
  \begin{tabular}{c c c c c c c c c c}
    \toprule
     & \multicolumn{4}{c}{YOLOv3} & \multicolumn{4}{c}{Mask-RCNN} \\
     \cmidrule(lr){2-5}\cmidrule(lr){6-9}
      Model & \phantom{....} ADE $(\downarrow)$ & \phantom{....} FDE $(\downarrow)$ & \phantom{....} AIOU $(\uparrow)$ & \phantom{....} FIOU $(\uparrow)$ & \phantom{....} ADE $(\downarrow)$ & \phantom{....} FDE $(\downarrow)$ & \phantom{....} AIOU $(\uparrow)$ & \phantom{....} FIOU $(\uparrow)$  \\
    \hline
    CV-CS & 32.9 & 60.5 & 51.4 & 26.7 & 31.6 & 57.6 & 46.0 & 21.3 \\
    LKF \cite{kalman} & 34.3 & 62.1 & 49.1 & 25.5 & 32.9 & 59.0 & 43.9 & 20.1 \\
    DTP \cite{dtp} & 28.7 & 52.4 & \textminus & \textminus & 26.7 & 48.5 & \textminus & \textminus\\
    FPL \cite{fpl} & 30.2 & 53.4 & \textminus & \textminus & 28.6 & 49.8 & \textminus & \textminus \\
    DTP-MOF & 29.0 & 52.2 & 54.6 & 30.8 & 27.3 & 49.2 & 49.6 & 25.1 \\
    FPL-MOF & 31.6 & 55.7 & 53.0 & 30.9 & 29.3 & 51.0 & 44.9 & 22.6 \\
    \textbf{STED} & \textbf{27.4} & \textbf{49.8} & \textbf{56.8} & \textbf{32.9} & \textbf{26.0} & \textbf{46.9} & \textbf{51.8} & \textbf{27.5}\\
    \bottomrule
  \end{tabular}\label{Citywalks-results}
  \end{center}
    \vspace{-0.2cm}
  \end{table*}

We adapt the following models for MOF, which are originally developed for trajectory forecasting. Each model is modified for full bounding box prediction assuming object scale is constant, or by adding additional output channels representing bounding box height and width for the learning-based approaches.

\textbf{Constant Velocity \& Constant Scale (CV-CS):} We adopt the simple constant velocity model, which is used widely as a baseline for trajectory forecasting models \cite{sociallstm,fpl,socialgan} and as a motion model for MOT \cite{online-dual-matching,quadruplet-networks,multi-target-multi-camera}. We use the previous 5 frames to compute the velocity, and find that using a constant scale performs better than linearly extrapolating a change in width and height.

\textbf{Linear Kalman Filter (LKF) \cite{kalman}:} The LKF is a widely-used method for tracking objects and predicting trajectories under noisy conditions. We use an LKF with initial parameters chosen using cross-validation and use the last updated motion value for forecasting. The LKF is one of the most popular motion models for MOT \cite{deepsort,candidatetracking,tracking-using-icf}.

\textbf{Future Person Localization (FPL) \cite{fpl}:} We adapt FPL, which uses pedestrian pose extracted using OpenPose \cite{openpose} and ego-motion estimation using optical flow extracted with FlowNet2 \cite{flownet2}.

\textbf{Dynamic Trajectory Predictor (DTP) \cite{dtp}:} We adapt DTP, which uses a CNN with past optical flow frames as input to predict future bounding boxes.

\subsection{Implementation details}

Clips from Citywalks are split into 3 folds, and the test set is further divided 50\% for validation and 50\% for testing for each fold. We use inter-city cross-validation, i.e., footage from cities in the validation/testing sets \textit{do not} appear in the training set. This challenging evaluation setup ensures that pedestrian identities from the training set do not appear at test time, and prevents models from overfitting to a particular environment.

\textbf{Bounding box feature encoder.} Bounding box vectors $B_t^i$ (defined in Section \ref{sec:bb-encoder}) are computed by taking the velocity of the object over the previous 5 timesteps, i.e., $v_t^x = x_t - x_{t-4}$ and $v_t^y = y_t - y_{t-4}$. Our feature encoder consists of a GRU with 512 hidden units which uses $B_{t-1}^i$ and the previous hidden state vector $h^e_{t-1}$ as input and outputs an updated hidden state vector $h^e_t$. We use GRUs rather than LSTMs as recurrent units in STED as we find the performance is similar while GRUs is less computationally demanding.

\textbf{Optical flow feature encoder.} We compute optical flow for each video frame using FlowNet2 \cite{flownet2}. The flow from within each pedestrian bounding box is then cropped, clipped to a range of $-50$ to $50$, scaled to a fixed size of $256 \times 256$, and normalized to a range of $0$ to $1$. We perform standard data augmentation, taking a random crop of size $224 \times 224$ and randomly horizontally flipping frames with probability 0.5 during training.  We train the optical flow feature encoder using ResNet50 \cite{resnet} as the backbone CNN architecture for 10k iterations with a batch size of 64 and learning rate of \num{1e-5} to predict future object locations as described in \cite{dtp} and then freeze the weights to use our flow encoder as a fixed feature extractor.

\textbf{Decoder.} As described in Section \ref{sec:decoder}, our decoder takes the concatenated feature vector $\phi_c$ as input. The decoder consists of another GRU with 512 hidden units. For each of the 60 timesteps to be predicted, the decoder takes $\phi_c$ and previous hidden state $h^d_{t-1}$ and outputs a new hidden state $h^d_t$. A linear layer takes the hidden state and generates a predicted bounding box for the respective timestep. The optical flow feature encoder is used as a fixed feature extractor, while the bounding box encoder and decoder are trained jointly end-to-end using an initial learning rate of \num{1e-3}, which is halved every 5 epochs. We use a batch size of 1024 and train the model for 20 epochs. The model is optimized using the smooth $\mathcal{L}_1$ loss, which we find to be more robust to outliers in the training data than the $\mathcal{L}_2$ loss.

\subsection{Results}

  \begin{figure*}[t]
     \centering
     \begin{subfigure}[t]{\textwidth}
         \centering
         \includegraphics[width=0.9\textwidth]{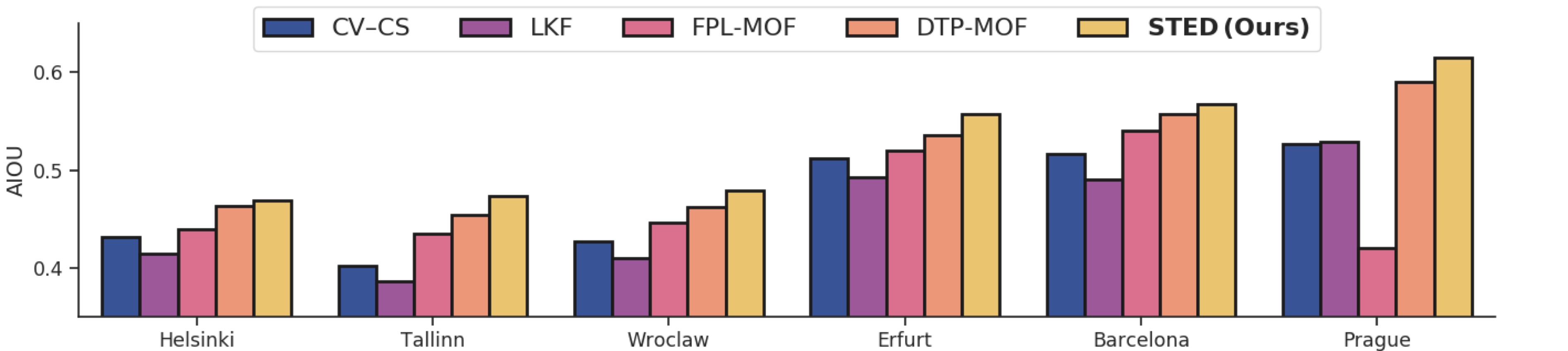}
         \caption{}
         \label{fig:iou-per-city}
        \vspace{-0.13cm}
     \end{subfigure}
     \begin{subfigure}[t]{0.3\textwidth}
         \centering
         \includegraphics[width=\textwidth]{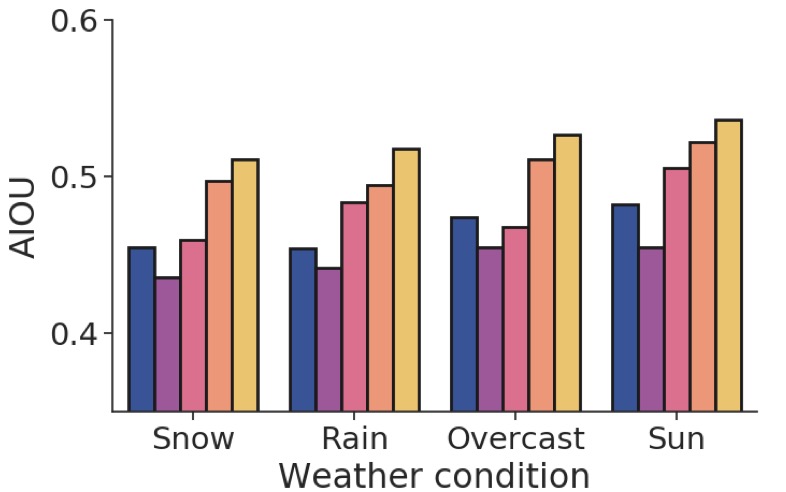}
         \caption{}
         \label{fig:iou-per-weather}
     \end{subfigure}
     \begin{subfigure}[t]{0.3\textwidth}
         \centering
         \includegraphics[width=\textwidth]{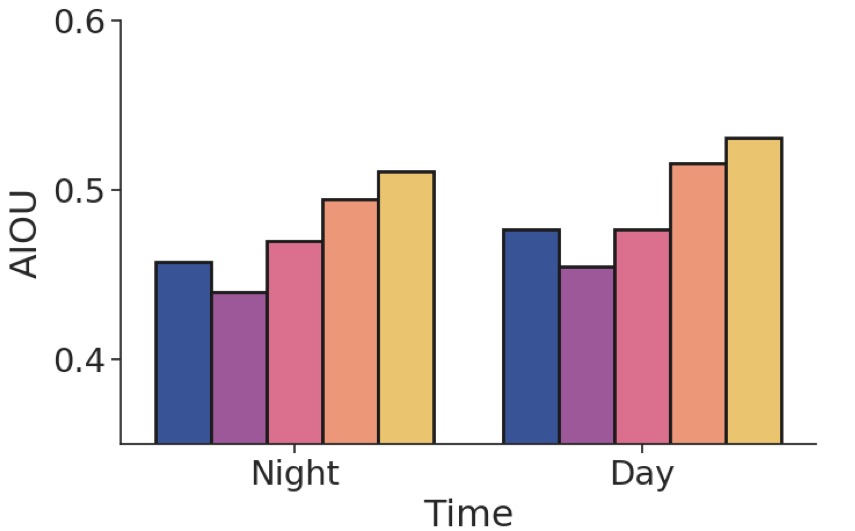}
         \caption{}
         \label{fig:iou-per-time}
     \end{subfigure}
     \begin{subfigure}[t]{0.3\textwidth}
         \centering
         \includegraphics[width=\textwidth]{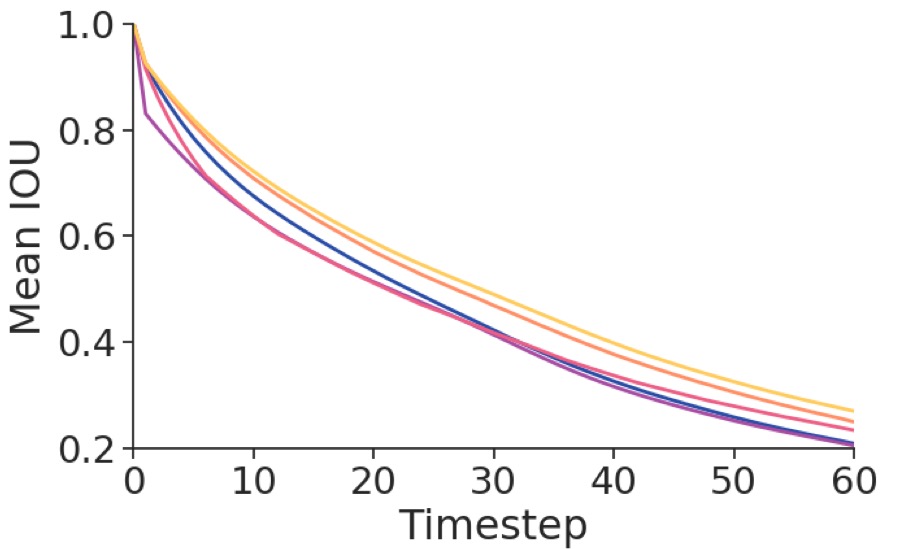}
         \caption{}
         \label{fig:iou-distribution}
     \end{subfigure}
         \vspace{-0.1cm}
        \caption{Performance analysis on Citywalks. Here, we report performance on both validation and test sets for all 3 folds to cover the entire dataset. Performance is broken down by (a) top 3 and bottom 3 cities by AIOU, (b) weather condition, (c) time of day and (d) future timestep.}
        \label{fig:perf-analysis}
\end{figure*}

  \begin{table}[t]
\setlength{\tabcolsep}{1pt} 
\begin{center}   \caption{Ablation study evaluating the bounding box (BB), optical flow (OF) encoders separately. Results are the mean of both annotation sets.}
  \begin{tabular}{ c  c  c}
    \toprule
    Model & \phantom{....} ADE / FDE $(\downarrow)$ & \phantom{....} AIOU / FIOU $(\uparrow)$  \\
    \hline
    BB-encoder & 29.6 / 53.2 & 51.5 / 27.9 \\
    OF-encoder & 27.5 / 50.0 & 53.2 / 28.8\\
    \textbf{Both encoders} & \textbf{26.7} / \textbf{48.4} & \textbf{54.3} / \textbf{30.2}\\
    \bottomrule
  \end{tabular}\label{tab:ablation}
  \end{center}
    \vspace{-0.6cm}
  \end{table}

We evaluate each model on the Citywalks dataset using both annotation sets and evaluate each component of STED separately. Finally, we evaluate the cross-dataset generalizability of each model on the MOT-17 dataset \cite{mot-16}.

\begin{figure*}
    \begin{center}
    \includegraphics[height=7.3cm,width=\linewidth]{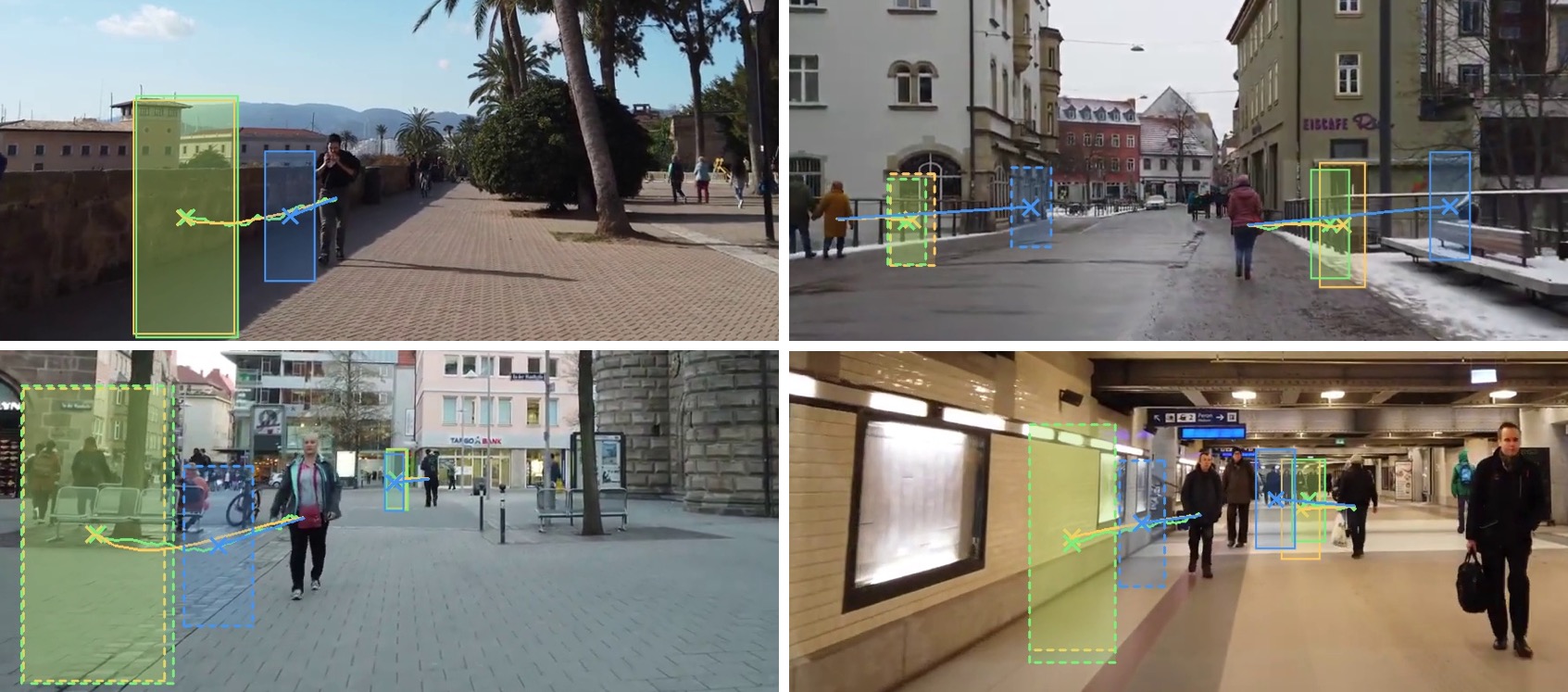}
    \end{center}
    \vspace{-0.5cm}
   \caption{Example successful object forecasts using our proposed model. Colours represent ground truth \textbf{(\textcolor{ground-truth}{Green})}, constant velocity and scale \textbf{(\textcolor{cv}{Blue})}, and STED \textbf{(\textcolor{ours}{Yellow})}. Forecasts are made for each of 60 timesteps in the future for all pedestrians in the scene, but here we visualize the predicted bounding box at $t = 60$ only and at most two pedestrians per frame for clarity. Line type (dashed/solid) denotes unique pedestrians. More example available at: \url{https://youtu.be/GPdNKE6fq6U}}
\label{fig:qual-success}
\end{figure*}

\begin{figure*}
    \begin{center}
    \includegraphics[width=\linewidth,height=7.3cm]{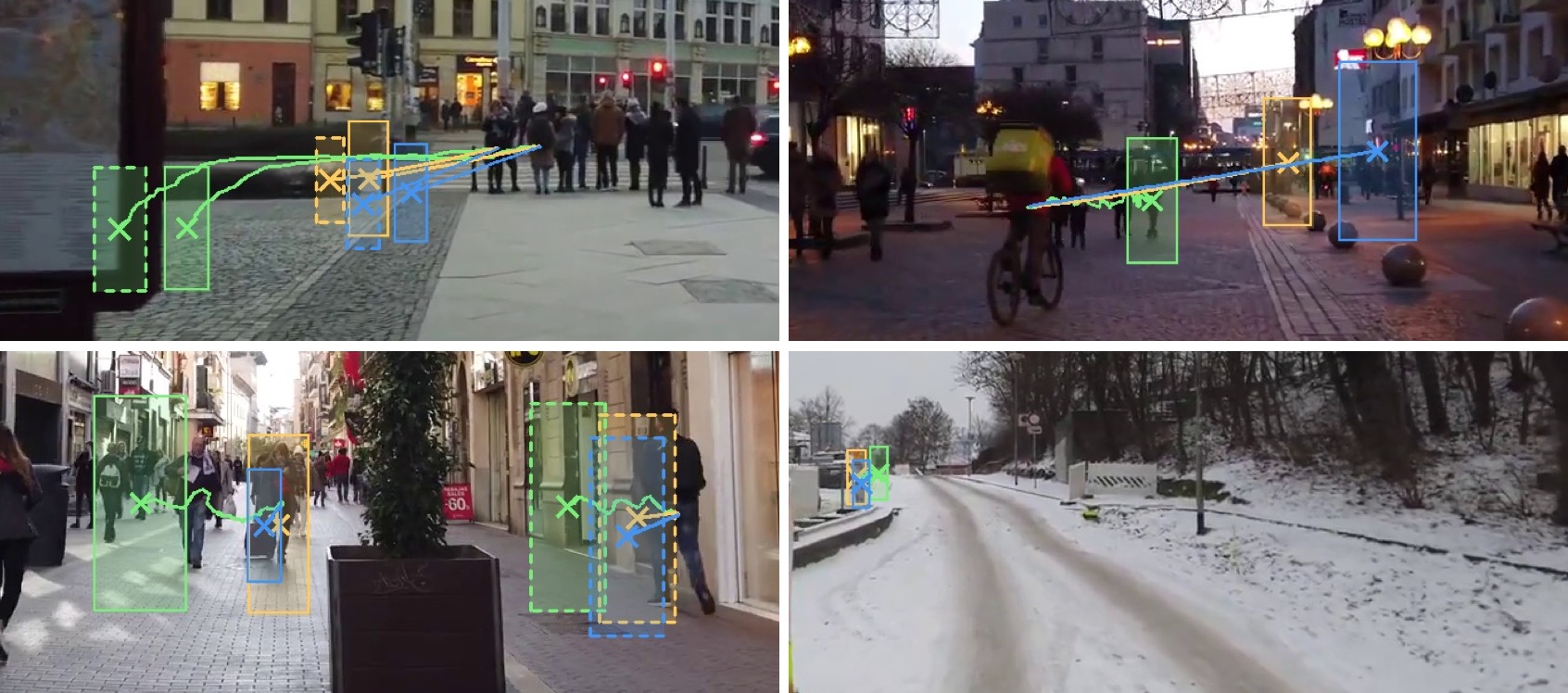}
    \end{center}
    \vspace{-0.5cm}
    \caption{Example unsuccessful object forecasts using our proposed model. Colours represent ground truth \textbf{(\textcolor{ground-truth}{Green})}, constant velocity and scale \textbf{(\textcolor{cv}{Blue})}, and STED \textbf{(\textcolor{ours}{Yellow})}. The examples highlight the difficulty of the Citywalks dataset, which contains several distant pedestrians and non-linear motions.}
    \vspace{-0.2cm}
\label{fig:qual-failure}
\end{figure*}

\textbf{Results on Citywalks.} Table \ref{tab:Citywalks-results} shows the ADE / FDE\footnote{A displacement of 50 pixels corresponds to 2.5\% of the total frame size at a resolution of $1280 \times 720$.} and AIOU / FIOU of all methods on Citywalks with both annotation sets. We evaluate the original DTP and FPL models for trajectory forecasting, as well as the versions modified for MOF. STED consistently performs better than existing approaches across all metrics, resulting in more precise bounding box forecasts. Fig. \ref{fig:qual-success} shows example bounding box predictions. STED implicitly anticipates both object and ego-motion in a diverse range of environments and situations. Fig. \ref{fig:qual-failure} shows failure cases. The model performs poorly in challenging conditions such as large ego-motions and when the pedestrian scale is small.

We further break down performance on Citywalks in Fig. \ref{fig:perf-analysis}.  We find that most models perform better for sequences recorded in cities with clear weather conditions (e.g., Barcelona, Prague) than, in particular, snow (e.g., Tallinn, Helsinki). To confirm this intuition, we further plot the performance in different weather conditions and at different times of the day. Finally, we plot the mean IOU at all predicted timesteps 1 to 60. The IOU of the predicted and ground-truth bounding boxes predictably declines quickly, particularly for earlier timesteps. STED maintains the best IOU throughout the full prediction horizon.

\textbf{Ablation study.} We evaluate the benefits of each component of our proposed model by evaluating them separately. Specifically, we use the bounding box encoder feature vector $\phi_b$ as input to the decoder, rather than the concatenated feature vector $\phi_c$. We repeat this for the optical flow encoder feature vector $\phi_f$. Table \ref{tab:ablation} show the results of our ablation study on Citywalks. Both the bounding box and optical flow encoders contribute to the overall performance.

\textbf{Computational complexity.} The most computationally expensive component of STED is computing optical flow. Our implementation uses FlowNet2, which requires 123ms to compute on an Nvidia GTX 1080 GPU \cite{flownet2}. This model may be replaced by more efficient methods, although we found the quality of optical flow to impact overall performance. Additional components, such as the CNN architecture or number of hidden units in the GRUs may be modified if real-time performance is required, at some cost in forecasting accuracy.

  \begin{table}[t]
\setlength{\tabcolsep}{1pt} 
\begin{center}   \caption{Results on MOT-17 after training on fold 3 of Citywalks. Models are not fine-tuned on MOT-17.}
    \vspace{-0.2cm}
  \begin{tabular}{ c  c  c }
    \toprule
    Model & \phantom{....} ADE / FDE $(\downarrow)$ & \phantom{....} AIOU / FIOU $(\uparrow)$  \\
    \hline
    CV-CS & \phantom{.} 58.9 / 104.7 & 43.8 / 21.5 \\
    LKF \cite{kalman} & \phantom{.} 62.0 / 110.2 & 41.6 / 20.1 \\
    FPL \cite{fpl} & 56.9 / 96.3 & \textminus \\
    DTP \cite{dtp} & 55.2 / 99.0 & \textminus \\
    FPL-MOF & 58.0 / 98.4 & 41.4 / 20.4\\
    \textbf{DTP-MOF} & 52.2 / 92.4 & \textbf{47.7} / \textbf{26.1}\\
    \textbf{STED} & \textbf{51.8} / \textbf{91.6} & 46.7  / 24.4 \\
    \bottomrule
  \end{tabular}\label{tab:mot-results}
  \end{center}
  \end{table}

\textbf{Cross-dataset evaluation.} In order to evaluate the generalizability of models trained on Citywalks, we use the popular MOT-17 dataset \cite{mot-16}. We use sequences 2, 9, 10, and 11 from the MOT-17 train set and discard sequences 4 and 13 as these sequences are filmed from an overhead perspective. We also discard sequence 5 due to the low image resolution and frame rate. We follow a similar pre-processing setup to Citywalks, discarding tracks shorter than 3 seconds. We also ensure pedestrians are occluded no more than 50\% of their total bounding box size using the annotations provided, resulting in 83 unique pedestrian tracks. We take each model trained on Citywalks and evaluate using each of the four sequences. Note that we do not modify the models and crucially we  \textit{do not} fine-tune on MOT-17. Table \ref{tab:mot-results} shows encouraging results suggesting that models trained on Citywalks generalize cross-dataset and to human-annotated bounding boxes. However, due to the small size of the MOT-17 dataset, these results should be treated with caution.

\section{Conclusion} \label{sec:conclusion}

We have introduced the task of multiple object forecasting and created the Citywalks dataset to facilitate future research. Crucially, we have shown that models trained on the Citywalks dataset can predict future object bounding boxes on the MOT-17 tracking benchmark more precisely than existing methods used by multiple object tracking. Our encoder-decoder model, STED, forecasts object bounding boxes up to 2 seconds in the future and anticipates non-linear motions. This development shows promise for building more sophisticated object forecasting models to aid object tracking in order to address common problems such as occlusions and missed detections.

\section*{Acknowledgements}
We would like to thank Shanaka Perera and Shuyang Sun for their insightful comments. This work is funded by the UK EPSRC (grant no. EP/L016400/1) and the EU Horizon 2020 project IDENTITY
(Project No. 690907). Our thanks to NVIDIA for supporting this research with their generous hardware donation. We would also like to thank Daniel Sczepansky for collecting the videos used in this research.

\clearpage
{\small
\bibliographystyle{ieee}
\bibliography{egbib}
}

\end{document}